\documentclass[11pt,a4paper]{article}
\usepackage[hyperref]{acl2019}
\usepackage{times}
\usepackage{latexsym}

\usepackage{url}

\aclfinalcopy 

\title{IITP at MEDIQA 2019: Systems Report for Natural Language Inference, Question Entailment and Question Answering}

\author{
Dibyanayan Bandyopadhyay$^1$ \quad Baban Gain$^1$ \quad Tanik Saikh$^{2}$ \quad Asif Ekbal$^{2}$\\
Government College Of Engineering And Textile Technology, Berhampore$^1$ \\ Indian Institute of Technology Patna$^2$ \\
\texttt{\{dibyanayan,gainbaban\}@gmail.com$^1$} \\
\texttt{\{1821cs08, asif\}@iitp.ac.in$^2$} \\
}

\date{}

\begin{document}
\maketitle
\begin{abstract}
This paper presents the experiments accomplished as a part of our participation in the MEDIQA challenge, an \cite{ACL-BioNLP} 
shared task. 
We participated in all the three tasks defined in this particular shared task. The tasks are \textit{viz. i. Natural Language Inference (NLI) ii. Recognizing Question Entailment(RQE)} and their application in medical \textit{Question Answering (QA)}. We submitted runs using multiple deep learning based systems (runs) for each of these three tasks. We submitted five system results in each of the NLI and RQE tasks, and four system results for the QA task. The systems yield encouraging results in all the three tasks.  The highest performance obtained in NLI, RQE and QA tasks are 81.8\%, 53.2\%, and 71.7\%, respectively.
\end{abstract}

\section{Introduction}
Natural Language Processing (NLP) in biomedical domain is an essential and challenging task. With the availability of the data in electronic form it is possible to apply Artificial intelligence (AI), machine learning and deep learning technologies to build data driven automated tools. These automated tools will be helpful in the field of medical science. An  ACL-BioNLP 2019 shared task, namely the MEDIQA challenge aims to attract further research efforts in NLI, RQE and their application in QA in medical domain. 
The motivation of this shared task is in a need to develop relevant methods, techniques, and gold standard data for inference and recognizing question entailment in medical domain and their application to improve domain specific Information Retrieval (IR) and Question Answering (QA) systems. The MEDIQA has defined several tasks related to \textit{Natural Language Inference, Question Entailment and Question Answering} in medical domain. We participated in all the three tasks defined in this shared task. We offer multiple systems for each the tasks. The workshop comprises of three tasks namely \textit{viz. i. Natural Language Inference (NLI): This task involves in identifying three inference relations between two sentences: i.e. Entailment, Neutral and Contradiction \cite{romanov-shivade-2018-lessons} ii. Recognizing Question Entailment (RQE): This task focuses on identifying entailment relation between two questions in the context of QA. The definition of question entailment is as follows: "a question A entails a question B if every answer to B is also a complete and/or partial answer to A" \cite{abacha2019question} and iii. Question Answering (QA): The goal of this task is to filter and improve the ranking of automatically retrieved answers. The existing medical QA system namely \textit{CHiQA} is applied to generate the input ranks. \cite{harabagiu-hickl-2006-methods,abacha2017overview,abacha2019question}}. We participated in all the three tasks defined above and submitted the results. Our proposed systems produce encouraging results.
\section{Proposed Method}
We propose multiple runs for each of the three tasks. The following subsections will discuss the methods applied to tackle each of these tasks. 
\subsection{Natural Language Inference}
In the task 1 the system has to decide the entailment relationship between a pair of texts i.e. either they are \textit{Entailment, Contradiction or Neutral}. The input to this task are sentence pairs and as output we wish to get the entailment relation between those two pieces of texts. We propose five runs for this task. The following set of hyper parameters are applied for the following runs. \textbf{Batch Size} = 32, \textbf{Learning Rate} = 2e-5, \textbf{Maximum Sequence Length} = 128, \textbf{number of epochs} = 10. The following points will discuss the approaches (i.e. runs). \\
\textit{\textbf{Run 1:}} Our first proposed method is based on a BioBERT \cite{lee2019biobert} model, i.e. a Bidirectional Encoder Representation from Transformer model pre-trained on biological data (both on Pub Med abstracts and PMC full-text articles). After getting the vector corresponding to the special classification token ([CLS]) from final hidden layer of this mode, we use it for classification. We use 2 dense (feed forward) layers and a softmax activation function at the end. Only the feed forward part is trained end to end for 10 epochs after getting output vector from BioBERT. In this method no fine tuning is used. This method yields an accuracy of 60.8\%.\\
\textit{\textbf{Run 2:}} The second approach is based on the Bidirectional Encoder Representation from Transformer (BERT) model (bert-base-uncased) \cite{devlin-etal-2019-bert}. We make use of this to get the embedding of the inputs to this system. Instead of using 2 feed forward layers at the end, we choose to use only 1 feed forward layer. The full model is then trained in an end to end manner. All of the parameters of BERT and the last feed forward layer are fine-tuned jointly for 10 epochs to maximize the log-probability of the correct label (Entailment, Neutral or Contradiction). This model produces an accuracy of 71.7\%. \\
\textit{\textbf{Run 3:}} 
We use a BioBERT model for this system. 
This BioBERT model is pre-trained on PubMed abstracts only. We apply only one feed forward layer at the end. The full model is fine tuned as described for the system in run 2. 
This method gives output with an accuracy of 77.1\%.\\
\textit{\textbf{Run 4:}} The system proposed in this run is same as the model of Run 3. The differences between them are as follows: 
\begin{itemize}
  \item The BioBERT model we used is a pre-trained model on both the PubMed abstracts and PMC full-text articles instead of only PubMed abstracts as in case of in run 3.
  \item Here, we combine the full dataset of MedNLI (14049 sentence pairs) for training. Whereas in the previous run we made use only 11232 sentence pairs for training.
\end{itemize}
Following these changes in run 4, the accuracy increases from 77.1\% in run 3 to 80.3\% in run 4.\\
\textit{\textbf{Run 5:}}
This model is the combination of three BioBERT models. Two of them are pre-trained on both the PubMed abstracts and PMC full text articles and the third one is pre-trained only on PubMed abstracts. We fine tune each of the models following the fine tuning process of run 4. We ensemble their predictions by voting each of them for a sentence pair. The label which gets the most vote is selected for final prediction. The accuracy increases to 81.8\%.\\
\subsection{Recognizing Question Entailment (RQE)}
Recognizing Question Entailment is an  important task. The objective of this task is to identify entailment between the two questions in the context of Question-Answering(QA). We use the following definition of question entailment: “a question A entails a question B if every answer to B is also a complete or partial answer to A”. 
We make use of the dataset provided by the task organizers'. We submit five runs which are broadly based on two approaches. The approaches are as follows: 
\begin{itemize}
  \item One is based on Siamese architecture \cite{mueller2016siamese}. This Siamese is based on the recurrent architectures for learning sentence similarity (Mueller et al., 2016). Here we feed the two questions (inputs) to two  Bidirectional Long Short Term Memory (Bi-LSTM) \cite{hochreiter1997long}, respectively. Both of their weights are initialized to the same. After obtaining the last hidden representations from both of these Bi-LSTMs, we concatenate them. This vector represents our input sentence pair. We feed this vector to a feed forward neural network layer. At the end, there is a softmax layer to perform a 2-class classification 
  (to Yes/No).
  \item In another one, we train (fine-tuned) a BioBERT model as described in the NLI task. We used the BioBERT model to perform a 3-way classification of a sentence pair into entailment, neutral or contradiction. The same approach is used here in RQE to classify a pair of questions into Yes or No. The hyper-parameters used in fine-tuning the BioBERT model are same as of task 1 (NLI) except here the training iteration is (i.e. epoch) 5. This is done so because the loss is decreasing rapidly between two training epochs, indicating over fitting on the train set.
\end{itemize}
Our proposed approaches are based on these methods with slight variations. The following points will show them.\\ 
\textit{\textbf{Run 1:}} Each question pair is having two questions (namely chq and faq). We assume the first question as the premise and the second one as the hypothesis. We extract these two questions from the training set. We obtain the vector representation of each word using Gensim Word2Vec \cite{rehurek_lrec}. The vector size is 50. Then vector representations of the words for both the question are fed to Siamese Network of Bidirectional LSTMs. We train the model with 50 epoch and achieve 53.2\% accuracy in test set.\\
\textit{\textbf{Run 2:}} 
In this method we make use of the BioBERT model. The model is pre-trained on PubMed abstracts and PMC full text articles. This task is essentially a sentence pair classification task. For each sentence pair we obtain a vector corresponding to ‘([CLS])’ token at the last layer.
The vector is subsequently fed into two dense layers followed by a final layer having softmax activation function layer. No fine-tuning is used here. We obtain an accuracy of 50.6\%. \\
\textit{\textbf{Run 3:}} 
Here also we use BioBERT, but it is pre-trained on PubMed abstracts only. We fine tune the model on the RQE training set consisting of 8588 pairs.
A feed forward layer with final layer with softmax activation is used at the end for 2-way classification. We obtain an accuracy of 48.1\%.\\
\textit{\textbf{Run 4:}} 
Instead of training word vector representation from scratch using Gensim Word2Vec, as in run 1, we obtain the vector representations of words from a trained Google News corpus (3 billion running words) word vector model (3 million 300-dimension English word vectors). The architecture is same as what is there in the run 1. We obtain an accuracy of 50.2\%.\\
\textit{\textbf{Run 5:}} We use a BioBERT model pre-trained on both PubMed abstracts and PMC full text articles. Then we fine tune on the RQE train set. Everything else is same as in run 3. The accuracy decreases to 48.9\%.\\

\subsection{Question Answering (QA)}
The objective of this Question Answering task is to filter and improve the ranking of automatically retrieved answers. The input ranks are generated by the existing medical QA system \textit{CHiQA}. We use BERT to predict the reference score between pairs and BM25 \cite{robertson2009probabilistic} to rank between them. First of all, the BERT is used as a sentence pair classifier model. The first token of every sequence is always the special classification token ([CLS]). The final hidden state (i.e., output of Transformer) corresponding to this token is used as the aggregate sequence representation for classification tasks. This final hidden state is a 768 dimensional vector (for bert-base) representing the input sentence pair. This vector is fed subsequently into one or more feed-forward layers with softmax activation function layer. We fine tune the whole system for 10 epochs to predict the reference score of test dataset. The predicted values of these reference score are subsequently used in the BM25 model. \textit{All the hyper parameters setting are same as in Task-1 except here a batch size of 28 is used because the maximum length of sequence is increased from 128 to 256.} It is to be noted that, it is too much memory consuming to train a BERT model with a batch size of 32 and a maximum sequence length of 256. We propose four runs to combat this problem.\\ 
\textit{\textbf{Pre-processing:}} The training dataset is divided into two files (QA-TrainingSet1-LiveQAMed and TrainingSet2-Alexa) both are containing 104 questions. They had 8.80 and 8.34 answers for every question on average, respectively. For a question with \textit{N} answers are converted to \textit{N} pairs with each pair containing the question, one of the answers, and their reference score.\\
\textit{\textbf{Run 1:}} 
All reference scores from training dataset are replaced by 1 if it is 3 or 4, and with 0 otherwise. The range of reference score (as given) is between 1 to 4 in dataset. 
Here we fine tune a BERT model (bert-base-uncased) with a feed forward layer at the end for 10 epochs to classify a sentence pair into 2 labels (0 or 1).
The trained model is then used to predict reference score of test set.
From the predicted result obtained from BERT, BM25 score for every question and their corresponding answers are calculated. 
All answers for a question whose predicted labels are 1 ('YES') are sorted in decreasing order of their BM25 scores. After that all the 'YES' labels are retrieved, and the same procedure is applied for all answers for the same question whose predicted label is 0 (NO). The obtained accuracy is 57.3\%.\\ 
\textit{\textbf{Run 2:}} All reference scores are kept intact, i.e. between 1 to 4. Here we use the BERT model (bert-base-uncased) and fine tune it on train set with a feed forward net at the end. 
From the predicted result obtained from BERT, pairs whose reference score is 4 or 3 are marked as 'YES' and whose reference score are 2 or 1 are marked as 'NO'.
BM25 score for every question and their corresponding answer is calculated. 
All answers for a question whose predicted label is 4 are sorted by decreasing order of their BM25 score. Same procedure is applied for all answers for the same question whose predicted label is 3,2 and 1, respectively. We obtain an accuracy of 65.1\% in this run.\\ 
\textit{\textbf{Run 3:}} 
Here the validation dataset is also included to the training set. We merged them. Instead of using a BERT model here we use BioBERT model which is pre-trained on PubMed abstracts and PMC full text articles. We fine tune this model as explained in run 1. The rest of the procedures are same as in run 2. The accuracy increases to 67.8\%.\\
\textit{\textbf{Run 4:}} This method is an ensemble of 5 BioBERT (PubMed-PMC) models and fine tuned on the train dataset. Each of the models is then evaluated on the validation set (which is included in training set of Run 3). It is seen that one of those models performs well than the ensemble of 5 models. The model is then used to predict reference score of the test set. The rest of the procedures is same as what is there in the Run 3. The accuracy is 71.7\% for this run.\\

\section{Experiments, Results and Discussions}
We submitted system results (runs) for all the three tasks. In all these tasks, we make use of the dataset released as a part of this shared task. In the following we 
discuss the dataset, evaluation results and the necessary analysis of the results obtained. \\ 
\textit{\textbf{Data:}} In the NLI task, the training and test instances are having 14049 and 405 number of sentence pairs, respectively. In task 2 (i.e. RQE), the training set is having 8588 number of pairs, out of which 4655 and 3933 pairs are having \textit{True} and \textit{False} class, respectively. The validation and test set are having 302 (true: 129 and false: 173) and 230 (true: 115 and false: 115) number of instances. In the QA task, training sets are provided from two domains \textit{viz. LiveQAMed and ii. Alexa}, each having 104 number of questions and at an average of 8.80 and 8.34 number of answers per question. There are 25 number of questions and at an average of 10.44 answers per question are there in the validation set. The test set for this task is having 150 question pairs and on an average 8.5 answer per question.

\textit{\textbf{Task 1(NLI):}} In the first task, we propose five runs. In all the tasks, we make use of either BERT or BioBERT models. We merge the input sentences pairs into a single sequence having maximum length of 128. They are separated by a special token ([SEP]). The first token of every sequence is always a special classification token ([CLS]). The final hidden state (i.e., output of Transformer) corresponding to this token is used as the aggregate sequence representation for the classification tasks. This final hidden state is a 768 dimensional vector (for bert-base) representing the input sentence pair. This vector is fed subsequently into one or more feed-forward layers with soft-max activation at the end for 3-way classification (Entailment, Neutral or Contradiction). The results for this task are shown in the Table \ref{resu-task1}.
\begin{table}[]
\centering
\begin{tabular}{|c|c|}
\hline
\textbf{Runs} & \textbf{Result(Accuracy(\%))} \\ \hline
1 & 60.8 \\ \hline
2 & 71.7 \\ \hline
3 & 77.1 \\ \hline
4 & 80.3 \\ \hline
5 & \textbf{81.8} \\ \hline
\end{tabular}
\caption{Submission results of all the five runs for the NLI task (Task-1)}
\label{resu-task1}
\end{table}
 We have discussed the way we can use a BERT model to perform sentence classification in medical domain. It is observed that an absolute improvement of 5.4\% in accuracy has been achieved by using a BioBERT (pre-trained on PubMed abstracts) model in run 3 instead of using the original BERT-base-uncased model (Pre-trained on Wikipedia and Book Corpus (as used in run 2)). The increase in result may be the effect of BioBERT, 
 because the other experimental set up remain same. 
 The reason for using 1 feed forward layer at the end of BERT models in all the runs except the run 1 (no fine tuning), using only one feed forward layer was putting the model into an under fitting state. While in case of fine tuning a large model, one feed forward is enough as suggested by \cite{devlin-etal-2019-bert}. Up to run 3, we make use of 11232 sentence pairs for the training. Those sentence pairs are same as the one used to train several models used in \cite{romanov-shivade-2018-lessons}. We use the remaining 2817 sentence pairs for validation. The validation set accuracy is always around 3-4\% higher than the test case accuracy for all the runs up to run 3. For getting the higher accuracy we combine all the 14049 pairs in the subsequent run. We get the accuracy of 81.8 \% which is the highest among all the proposed methods. As per our knowledge, in the official results of NLI task we stand at 12th position among the 17 official teams which participated for the NLI task.\\ 

\textit{\textbf{Task 2 (RQE):}} In the second task i.e. task of Recognising Question Entailment, we propose five runs. The results are shown in the Table \ref{resu-task2}.
\begin{table}[]
\centering
\begin{tabular}{|c|c|}
\hline
\textbf{Runs} & \textbf{Result(Accuracy(\%))} \\ \hline
1 & \textbf{53.2} \\ \hline
2 & 50.6 \\ \hline
3 & 48.1 \\ \hline
4 & 50.2 \\ \hline
5 & 48.9 \\ \hline
\end{tabular}
\caption{Submission results in all the five runs for the RQE Task (Task-2)}
\label{resu-task2}
\end{table}
It is interesting to note the variation in accuracy for the different runs. Siamese architecture performs much better here. Another peculiarity is that fine tuning BERT hurts the performance while using pre-trained BERT embedding without fine tuning seems to be more useful. This is concluded by observing the results of run 2 and run 3. In run 2, we used only pre-trained BERT embedding for ([CLS]) token for classification, whereas in run 3, we fine tuned the BERT model. The highest accuracy is achieved by a Siamese Model consisting of 2 Bi-LSTMs with shared weights and a dense layers. In this task, 12 teams submitted their systems, and we stood the 10th position. \\ 

\textit{\textbf{Task 3 (QA):}} In this task, we offer 4 runs to tackle the problem. The results for this are shown in the Table \ref{resu-task3}. 
\begin{table}[]
\scriptsize
\centering
\begin{tabular}{|c|c|c|c|c|}
\hline
 & \multicolumn{4}{c|}{Results} \\ \hline
\textbf{Runs} & \textbf{Accuracy(\%)} & \textbf{Spearman’s Rho} & \textbf{MRR} & \textbf{Precision} \\ \hline
1 & 57.3 & 0.053 & 0.8241 & 0.5610 \\ \hline
2 & 65.1 & 0.042 & 0.7811 & 0.7235 \\ \hline
3 & 67.8 & 0.034 & 0.8366 & 0.7421 \\ \hline
4 & \textbf{71.7} & 0.024 & 0.8611 & 0.7936 \\ \hline
\end{tabular}
\caption{Results obtained in all the four runs for the QA Task (Task - 3), where, MRR: Mean Reciprocal Rank}
\label{resu-task3}
\end{table}
As we can see from the above discussions, the systems we build for this task comprises of two components, they are BERT and BM25. The BERT is used to predict the reference score of the test dataset. We rank the predicted scores using BM25. The BM25 part of the system is same for all the runs. In this task, participants are encouraged to compute the Mean Reciprocal Rank (MRR), Precision, and Spearman's Rank Correlation Coefficient as the evaluation measures in addition to Accuracy. We actually used BioBERT instead of original BERT from the run 3, which increases the accuracy with an absolute margin of 2.7\% (65.1 to 67.8\%). Using BioBERT we observe an improvement in MRR by 5.5\%. Our best run with an accuracy of 71.7\% attains the position of 6th among 10 teams in the official result.\\

\section{Conclusion and Future Work}
In this paper, we present our system details and the results of various runs that reported as a part of our participation in the MEDIQA challenge. 
In this shared task three tasks, namely \textit{viz. i. Natural Language Inference ii. Question Entailment and iii. Question Answering} were introduced in the medical domain. 
We offer multiple systems (runs) for each of these tasks. Most of the proposed models are based on BERT/Bio-BERT embedding and BM25. These models yields encouraging performance in all the tasks.
In future we would like to extend our work as follows:
\begin{itemize}
  \item Detailed analysis of the top-scoring models to understand their techniques and findings. 
  \item We can do the task of NLI by fostering an \textit{Embedding from Language model (EMLo)} based model and do a comparative analysis with BERT based model.
\end{itemize}
\bibliography{acl2019}

\begin{thebibliography}{11}
\expandafter\ifx\csname natexlab\endcsname\relax\def\natexlab#1{#1}\fi

\bibitem[{Abacha et~al.(2017)Abacha, Agichtein, Pinter, and
  Demner-Fushman}]{abacha2017overview}
Asma~Ben Abacha, Eugene Agichtein, Yuval Pinter, and Dina Demner-Fushman. 2017.
\newblock Overview of the medical question answering task at {TREC} 2017
  {LiveQA}.
\newblock In \emph{Proceedings of The Twenty-Sixth Text REtrieval Conference,
  TREC}, pages 15--17.

\bibitem[{Abacha and Demner-Fushman(2019)}]{abacha2019question}
Asma~Ben Abacha and Dina Demner-Fushman. 2019.
\newblock A question-entailment approach to question answering.
\newblock \emph{arXiv preprint arXiv:1901.08079}.

\bibitem[{Abacha et~al.(2019)Abacha, Shivade, and Demner-Fushman}]{ACL-BioNLP}
Asma~Ben Abacha, Chaitanya Shivade, and Dina Demner-Fushman. 2019.
\newblock Overview of the {MediQA} 2019 shared task on textual inference,
  question entailment and question answering.
\newblock In \emph{Proceedings of the BioNLP 2019 workshop, Florence, Italy,
  August 1, 2019}. Association for Computational Linguistics.

\bibitem[{Devlin et~al.(2019)Devlin, Chang, Lee, and
  Toutanova}]{devlin-etal-2019-bert}
Jacob Devlin, Ming-Wei Chang, Kenton Lee, and Kristina Toutanova. 2019.
\newblock {BERT}: Pre-training of deep bidirectional transformers for language
  understanding.
\newblock In \emph{Proceedings of the 2019 Conference of the North {A}merican
  Chapter of the Association for Computational Linguistics: Human Language
  Technologies, Volume 1 (Long and Short Papers)}, pages 4171--4186,
  Minneapolis, Minnesota. Association for Computational Linguistics.

\bibitem[{Harabagiu and Hickl(2006)}]{harabagiu-hickl-2006-methods}
Sanda Harabagiu and Andrew Hickl. 2006.
\newblock Methods for using textual entailment in open-domain question
  answering.
\newblock In \emph{Proceedings of the 21st International Conference on
  Computational Linguistics and 44th Annual Meeting of the Association for
  Computational Linguistics}, pages 905--912, Sydney, Australia. Association
  for Computational Linguistics.

\bibitem[{Hochreiter and Schmidhuber(1997)}]{hochreiter1997long}
Sepp Hochreiter and J{\"u}rgen Schmidhuber. 1997.
\newblock Long short-term memory.
\newblock \emph{Neural computation}, 9(8):1735--1780.

\bibitem[{Lee et~al.(2019)Lee, Yoon, Kim, Kim, Kim, So, and
  Kang}]{lee2019biobert}
Jinhyuk Lee, Wonjin Yoon, Sungdong Kim, Donghyeon Kim, Sunkyu Kim, Chan~Ho So,
  and Jaewoo Kang. 2019.
\newblock Bio{BERT}: pre-trained biomedical language representation model for
  biomedical text mining.
\newblock \emph{arXiv preprint arXiv:1901.08746}.

\bibitem[{Mueller and Thyagarajan(2016)}]{mueller2016siamese}
Jonas Mueller and Aditya Thyagarajan. 2016.
\newblock Siamese recurrent architectures for learning sentence similarity.
\newblock In \emph{Thirtieth AAAI Conference on Artificial Intelligence}.

\bibitem[{{\v R}eh{\r u}{\v r}ek and Sojka(2010)}]{rehurek_lrec}
Radim {\v R}eh{\r u}{\v r}ek and Petr Sojka. 2010.
\newblock {Software Framework for Topic Modelling with Large Corpora}.
\newblock In \emph{{Proceedings of the LREC 2010 Workshop on New Challenges for
  NLP Frameworks}}, pages 45--50, Valletta, Malta. ELRA.

\bibitem[{Robertson et~al.(2009)Robertson, Zaragoza
  et~al.}]{robertson2009probabilistic}
Stephen Robertson, Hugo Zaragoza, et~al. 2009.
\newblock The probabilistic relevance framework: {BM25} and beyond.
\newblock \emph{Foundations and Trends{\textregistered} in Information
  Retrieval}, 3(4):333--389.

\bibitem[{Romanov and Shivade(2018)}]{romanov-shivade-2018-lessons}
Alexey Romanov and Chaitanya Shivade. 2018.
\newblock Lessons from natural language inference in the clinical domain.
\newblock In \emph{Proceedings of the 2018 Conference on Empirical Methods in
  Natural Language Processing}, pages 1586--1596, Brussels, Belgium.
  Association for Computational Linguistics.

\end{thebibliography}
\bibliographystyle{acl_natbib}

\appendix

\end{document}